\documentclass[10pt, a4paper]{article}

\usepackage{makecell}
\usepackage{hyperref}
\usepackage{graphicx}
\usepackage{multirow}
\usepackage{amsmath}
\usepackage{xcolor}
\usepackage{booktabs}
\usepackage{caption}
\usepackage{subcaption}
\usepackage{amsfonts}
\usepackage{fixltx2e}
\usepackage{tabularx}
\usepackage{microtype}

\usepackage{inconsolata}
\usepackage{amssymb}% http://ctan.org/pkg/amssymb
\usepackage{pifont}% http://ctan.org/pkg/pifont
\newcommand{\cmark}{\ding{51}}%
\newcommand{\xmark}{\ding{55}}%

\usepackage[]{lrec-coling2024} % this is the new style

\title{DMON: A Simple yet Effective Approach for Argument Structure Learning}

\name{Wei Sun, Mingxiao Li, Jingyuan Sun, Jesse Davis, Marie-Francine Moens} 

\address{Department of Computer Science, KU Leuven \\
         Celestijnenlaan 200A 3001 Heverlee, Belgium \\
         {Sun.Wei, Mingxiao.Li, Jingyuan.Sun, Jesse.Davis,Sien.Moens}@kuleuven.be\\
         }

\abstract{
Argument structure learning~(ASL) entails predicting relations between arguments. 
Because it can structure a document to facilitate its understanding, it has been widely applied in many fields~(medical, commercial, and scientific domains).  
Despite its broad utilization, ASL remains a challenging task because it involves examining the complex relationships between the sentences in a potentially unstructured discourse. 
To resolve this problem, we have developed a simple yet effective approach called Dual-tower Multi-scale cOnvolution neural Network~(DMON) for the ASL task.
Specifically, we organize arguments into a relationship matrix that together with the argument embeddings forms a relationship tensor and design a mechanism to capture relations with contextual arguments.
Experimental results on three different-domain argument mining datasets demonstrate that our framework  outperforms state-of-the-art models.
The code is available at \url{https://github.com/VRCMF/DMON.git}. 
 \\ \newline  }

\begin{document}

\maketitleabstract

\section{Introduction}

% \jesse{i've rewritten this a bit}

\begin{figure}[t]
\centering
\includegraphics[width=0.9\linewidth]{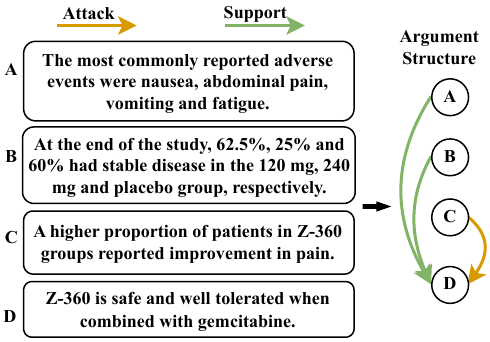}
\caption{A simple example of an argumentative structure showing attack (orange arrow) and support (green arrow) relationships.} %Other un-tagged relations are labeled as `no-relation'.
%Argument Structure Learning is to predict labels for all relations between each argument. }
\captionsetup{justification=centering}
\label{fig:what}
\end{figure}

{Argument structure learning~(ASL)~\cite{Moens2013, lawrence2020argument} involves detecting and tagging relationships between argumentative components in a text. 
Figure~\ref{fig:what} shows an illustrative example  of an argumentative structure for a medical report where pairs of sentences are annotated with whether there is a supportive or attacking relationship between them.  
This problem is a cornerstone in the semantic analysis of natural language text because it helps to elucidate the relational structure. 
Consequently, it helps facilitate more accurate and deeper comprehension of text and hence plays a critical role in various NLP applications such as patient-generated content analysis~\cite{mayer2020transformer,stylianou2021transformed}, legal reasoning~\cite{poudyal2020echr}, and opinion mining~\cite{niculae2017argument}.}  
%Figure~\ref{fig:what} shows an example of an argumentative structure for a medical report. 

%Argument structure learning~(ASL)~\cite{Moens2013}, detecting and tagging relationships between argumentative components in a text, is a cornerstone in the semantic analysis of natural language text.  It plays a critical role in various NLP applications such as patient-generated content analysis~\cite{mayer2020transformer}, legal reasoning~\cite{poudyal2020echr}, and opinion mining~\cite{niculae2017argument}  by helping to understand the relational structure, thereby facilitating more accurate and deep comprehension of the text. Figure~\ref{fig:what} shows an example of an argumentative structure for a medical report. 

% \jesse{For me, the following paragraph does not tell a coherent story. Are you trying to list challenges? If so, I'd enumerate the challenges and in particular the ones addressed in this paper. Otherwise I think the main argument is that people ignore context. I've written something that follows the original paragraph (which directly follow's this point) focusing on the last point. }

Despite of its broad application, solving the ASL task is still challenging due to the complexity of text structures and diversity of relationships. 
Moreover, real-world data often contains inconsistencies and are largely unstructured. 
Fully understanding the relationship between two arguments often require contextual knowledge from other arguments, or even their relationships.  
%\citet{mayer2020transformer, stylianou2021transformed, galassi2021multi} tried to conduct pairwise relation classification for ASL without contextual information, yielding sub-optimal classification performance.
%
%The challenge of ASL is further complicated by the scarcity of annotated data.

{ A key challenge posed by ASL is that fully understanding the relationship between two arguments often requires capturing contextual knowledge about other arguments and their relationships.} 
% Contextual relationships represent incoming and outcoming edges for a given arguments. 
% \jesse{JESSE: Here give an example of context}. 
In Figure~\ref{fig:what}, to classify the relationship $\textbf{C} \rightarrow \textbf{D}$, 
%the contextual arguments are $\textbf{B}$ and $\textbf{C}$, and the 
% contextual argument relationships are, for instance, relationships between \textbf{C} and other arguments in the discourse. 
examples of contextual argument relationships are $\textbf{A} \rightarrow \textbf{D}$ and $\textbf{B} \rightarrow \textbf{D}$.
%Initially, work tried to classify pairwise relations without contextual information but obtained sub-optimal performance~\cite{mayer2020transformer, stylianou2021transformed, galassi2021multi}. 
\citet{mayer2020transformer}, \citet{stylianou2021transformed} and \citet{galassi2021multi} tried to conduct pairwise relation classification for ASL without contextual information, yielding sub-optimal classification performance.
A more recent attempt by \citet{hua2022efficient}
encodes the contextual arguments with a transformer architecture. 
This helped to improve its accuracy, but they still ignored the relationships between contextual arguments.

% \jesse{The graph part needs to be highlighted in the below text.}

\begin{figure}[h]
\centering
\includegraphics[width=0.65\linewidth]{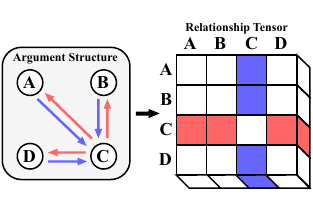}
\caption{We select argument $C$ as the observation object. 
This example shows the correlation between the head~(red) and tail~(blue) relationship information and a relationship tensor. 
Each element in this relationship tensor is the concatenation of two arguments. }
\captionsetup{justification=centering}
\label{fig:bidirection}
\end{figure}

{In this paper, we for the first time propose to exploit contextual argument relationships to solve the ASL task. 
As shown in Figure~\ref{fig:bidirection}, we represent the argument structure as a relationship tensor to capture the contextual information about argument relationships. 
This also allows us to naturally model the relationships between pairs of arguments that can be bidirectional and asymmetric. 
%To account for these facts, w
We propose a bidirectional learning approach that uses a separate model for each direction.  
Moreover, training is also hampered by the fact that there is limited labeled data for ASL problems due to the high annotation costs. 
Therefore, we propose a cropping strategy that randomly samples a subtensor that maintains the ordering of the selected relationships.}

% To implement this idea on the representation learning,
% ---------------------------
% ASL datasets suffer from the data scarcity problem, i.e., the number of training graphs is limited, because of the high cost of annotating argumentative graphs.
% We developed a cropping strategy to alleviate this problem. 
% % In each training iteration, %During training a %symmetry 
% A cropping strategy samples an argumentative sub-matrix from an relationship matrix at each training iteration. 
% Rather than randomly sampling argument relationships from the relationship matrix, the 
% %symmetry 
% cropped sub-matrix keeps the vertical and horizontal ordering of the selected relationships.
% As shown in our experimental results and their analysis, the cropping will improve generalization when training with limited data.  
% ---------------------------

% willanxywc: This paragraph is largely abundant. The method descriptions shoud mostly take one paragraph in theintroduction.
% willanxywc: Pleasev try to merge the two paragraphs of method descriptions into one

The main contributions of this paper are the following: 
\begin{itemize}
    % \item We propose an alternative approach to conduct argument structure learning by connecting argument structure and 
    % \item We propose a novel approach to easily encode contextual arguments and relationships into an argumentative tensor.
    % %by connecting argument structure with the relationship matrix. 
    % \item We develop a bidirectional learning mechanism to obtain head and tail information for detecting and classifying argument relationships.
    % \item The head and tail information is naturally contained within relationship matrices. 
    % Experimental results state the superior of transforming arguments into an relationship matrix. 
    \item We propose a novel approach called DMON to encode contextual arguments and their relationships by connecting the argument structure with a relationship tensor. 
    \item We propose a bidirectional learning mechanism that allows distinguishing head and tail arguments in a relationship. 
    %and two kinds of neighboring relation information to benefit the ASL task. 
    \item We design a %symmetry 
    cropping strategy to handle the scarcity of training data.
    \item Experimental results on three different-domain argument mining datasets show that our method outperforms state-of-the-art models for the ASL task.
\end{itemize}

First, we discuss related works in section~\ref{sec:related}.
Next, in section~\ref{sec:methods}, we provide a detailed description of the DMON framework. 
Then, in section~\ref{sec:exp}, we present experimental results on two argument mining datasets from different domains and conduct an ablation study to analyze the proposed framework and its outcomes. 
Finally, in section~\ref{sec:conclude}, we summarize this paper.

\begin{figure*}[!ht]
\centering
\includegraphics[width=\linewidth]{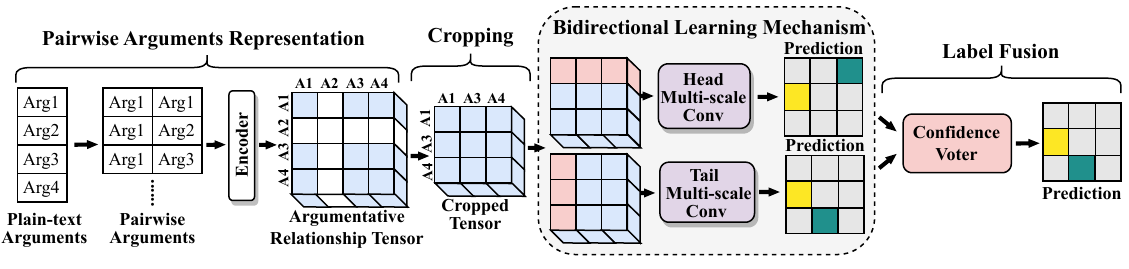}
\caption{The structure of Multi-scale Residual Convolution Neural Network~(DMON) during training. 
During testing, no cropping mechanism is used.
In this example, the yellow and green cells in a prediction matrix correspond to attack and support relations.)}
\captionsetup{justification=centering}
\label{fig:arch}
\end{figure*}

\section{Related Work}
\label{sec:related}

\subsection{Argument Structure Learning}
Argument Structure Learning is a challenging but essential task in text mining~\cite{Moens2013,lawrence2020argument}. 
% \jesse{This is repeated almost verbatim from the intro, so I'd drop it here.} It holds significant potential for various applications including medical, legal and commercial fields. 
% % , such as patient-generated content analysis~\cite{mayer2020transformer}, legal reasoning~\cite{poudyal2020echr}, and opinion mining~\cite{niculae2017argument}. 
% Argument mining involves two main tasks: argument identification and argument relation classification. 
% Prior works~\cite{galassi2021multi,mayer2020transformer,stylianou2021transformed,bao2021neural} achieved competitive performance on argument identification task, but accurately detecting the argumentative relationships remains a challenge. 
%Most papers perform pairwise ASL and classify two concatenated sentences   to deal with the argument relation classification because of it simplicity and effectiveness.
% \jesse{Most papers tackle argument relation classification by performing pairwise ASL and classifying two concatenated sentences due to it simplicity and effectiveness.}
Most papers tackle argument relationship classification by performing pairwise ASL and classifying two concatenated sentences due to it simplicity and effectiveness \cite{mayer2020transformer, stylianou2021transformed, galassi2021multi}.

%Encoding contextual arguments and relations is crucial for effective argument structure learning. 
In \citet{hua2022efficient}, argument components and their contextual sentences were encoded together using a RoBERTa encoder to obtain neighboring arguments information. 
The experimental results demonstrate that incorporating contextual information enhances performance. 
%on the ASL task. 
This method primarily relies on self-attention to capture relationships between the concatenated sentences. 
%argument structure. 
However, self-attention can be computationally expensive as it involves computing a pairwise similarity matrix for every token in all argument components. 
% Moreover, this approach may not capture the entire argument structure since it does not consider each argument holistically, resulting in a limited understanding of the relationships between different components of the argument. 
% Therefore, the challenge lies in developing an efficient approach that incorporates the broader argument context and leverages contextual information to improve argument structure learning.
In this paper we explicitly model contextual argumentative relationships and, given that a convolution operation can effectively capture input-data features~\cite{alzubaidi2021review, andreoli2019convolution, dumoulin2016guide}, we utilize convolutional modules to represent a pair of arguments and contextual pairs of arguments. 
%capture the structure of input arguments. As a result, we can encode information of contextual argumentative relationships
%Meanwhile, DMON can also encode contextual arguments as each cell contains information about them. 
% Our work is the first one to encode neighboring relations to benefit the ASL task. 

\subsection{%relationship Matrix
Structured Learning in NLP
}
Structured learning~(SL) also called structured prediction plays a crucial role in many NLP tasks.
It models complex relationships and dependencies within text data to improve the performance of various discourse-related applications, such as sentiment analysis~\cite{ein2022fortunately}, and summarization~\cite{xu2019discourse}.
% and machine translation~\cite{jiang2023discourse,xiong2019modeling}.
%For example, discourse structure can be used in text summarization to help identify the most important and relevant parts of a text, as well as to structure and organize the summary in a coherent and logical way.
%Discourse structure can be used in sentiment analysis to help understand the context, structure, and meaning of a text, as well as to capture the nuances and variations of sentiment across different segments or levels of a text. 
In argument mining, \citet{niculae2017argument} use a structured support vector machine, while \citet{bao2021neural} implement transition-based dependency parsing to reveal the argumentative structure.
% For example, in discourse parsing, CRFs can be used to predict the discourse structure of a document by considering the relationships between sentences, identifying discourse relations, and labeling discourse elements.
%An relationship matrix is a square matrix employed to represent a finite graph. 
%Its elements indicate the relationship between vertices~(arguments in our case). 
%Given that an argumentative graph is a directed acyclic graph~(DAG), the argument structure graph can be represented as an asymmetric relationship matrix.
%To the best of our knowledge, this paper is the pioneering work that represents the argument structure using an relationship matrix to capture neighboring relation information. 

\section{Method}
\label{sec:methods}
% The goal of the ASL is to predict relations between arguments. 
\noindent \textbf{Problem Setting:} %Let us consider a given document containing $n$ sentences as potential arguments, which we will regard it as the argument set $A$. 
%Given a document containing $n$ sentences, we treat each one as a potential argument and use $A$ to denote the argument set.  We convert $A$ into the set of ordered pairs $C_A = \{(a_i, a_j) | a_i \in A~\text{and}~a_i \in A \} \in \mathbb{R}^{n^2}$
{Given a document containing $n$ sentences, we treat each one as a potential argument and use $A$ to denote the argument set.  We convert $A$ into the set of ordered pairs $C_A = \{(a_i, a_j) | a_j \in A~\text{and}~a_i \in A \} \in \mathbb{R}^{n^2}$.}
%We transform the argument set into the set of ordered pairs, which is denoted as $C_A = \{(a_i, a_j) | a_i \in A~\text{and}~a_i \in A \} \in \mathbb{R}^{n^2}$.
The goal of ASL is to classify all relations in the set $C_A$ with the domain-specific labels contained in the given dataset. 

\noindent \textbf{Method Overview:} We introduce a Dual-tower Multi-scale cOnvolution neural Network~(DMON) for the ASL task. 
%The training of our
The model has four components   
%involves four main steps 
(Figure~\ref{fig:arch}). 
% Firstly, we transform the original arguments into pairwise arguments and generate hidden representations of the pairwise arguments by an encoder. 
Firstly, we use an encoder to extract pairwise argument representations. 
%We unflatten the representations and organize them with the help of an relationship matrix representation.
%, which organizes argumentative relationship matrices. 
% \jesse{We unflatten the representations and organize them using an relationship matrix. An relationship matrix is a square matrix where each index corresponds to a vertex in the graph.  Each entry indicates whether an edge connects the corresponds vertices~(arguments in our case).}
%{We unflatten the representations and organize them using an relationship matrix. An relationship matrix is a square matrix where each index corresponds to a vertex in the graph.  Each entry indicates whether an edge connects the corresponds vertices~(arguments in our case).}
Given that an argumentative graph is a directed acyclic graph~(DAG), potential argumentative relationships can be represented as an asymmetric relationship matrix, or as a relationship tensor when it includes the pairwise argument embeddings.
%To the best of our knowledge, this paper is the pioneering work that represents the argument structure using an relationship matrix to capture neighboring relation information. 
Secondly, during training, a %alignment 
cropping strategy {selects} sub-tensors from the relationship tensor. 
% \jesse{This helps alleviate the scarcity of training data scarcity and increases the generalization capability of the trained model.}
%{This helps alleviate the scarcity of training data scarcity and increases the generalization capability of the trained model.}
Thirdly, a bidirectional learning mechanism is applied to the cropped relationship tensors to capture contextual arguments and their relationships. %information. 
% as the representation of the feature matrices can change depending on the direction of convolution. 
% as head and tail information can be obtained by using horizontal and vertical multi-scale convolution, respectively. 
% Two separated branches of the bidirectional learning mechanism is to resolve head-tail information confusion problem and the multi-scale convolution deals with variant context window problem.
Finally, we employ
%\jesse{employ} \jesse{if this is novel, you can a "we employ a novel label fusion strategy"} 
label fusion to merge two predicted label matrices into one label matrix.
% Finally, label fusion involves merging two label matrices that are generated by the bidirectional learning mechanism to provide final label matrix. 
%Figure~\ref{fig:arch} shows the overall architecture of our proposed framework.
During testing, we feed full relationship tensors instead of cropped tensors into the model.

\subsection{Pairwise Arguments Representation}
% \jesse{Jesse's new text: 
{Following the literature~\cite{mayer2020transformer,hua2022efficient,stylianou2021transformed}, we treat each sentence as a potential argument. 
To capture pairwise interactions, we create all possible pairs $(a_i,a_j)$ and concatenate them with a special token placed between them.  %separated by a special token.
{Next, we use a linkBERT model to encode the paired arguments into an average pooled embedding with dimensionality $d$}. 
We organize these into a tensor $\mathbf{H} \in \mathbb{R}^{n \times n \times d}$.}  %\jesse{jesse: now explain in your notation the orientation?}
Figure~\ref{fig:bidirection} illustrates how to transform the sentences of a discourse into a relationship tensor. 
% Each cell in the relationship tensor is the representation of a concatenated paired arguments, which represents incoming or outcoming edges for a given argument. 
% \jesse{Jesse: I still do not get the point about the graph and how the graph structure, i.e., edges, interacts with the tensor.}
Each cell in the relationship tensor represents the concatenation of a coupled argument, which typically consists of two elements: the first element is known as the head argument, while the second element is referred to as the tail argument.
\subsection{Cropping Strategy}

We use a cropping strategy to mitigate scarcity of labeled training data. 
During each training iteration, we sample a new sub-tensor $\mathbf{H}^\prime \in \mathbb{R}^{m \times m \times d}$ from $\mathbf{H}$.  
Concretely, we sample $m$ indices  $\{i_0, i_1, \ldots, i_{m-1}\}$ without replacement from $\{0,1,\ldots,n-1\}$ where $m < n$ and $m$ is called the window size.
%This induces a sub-tensor $\mathbf{H} \in \mathbb{R}^{m \times m \times d}$.
%where...
Now we describe in mathematical notation how these indices induce a sub-tensor.
% , maybe it should be $H^{'}_{j,k} = H_{i_j,i_k}$. 
$\mathbf{H}^{'}_{j,k}$ represents an element of sub-tensor $\mathbf{H}^\prime$ whose row and column indexes are $j$ and $k$, and $\mathbf{H}^\prime_{j,k} = \mathbf{H}_{i_j,i_k}$.
 The cropping strategy keeps the discourse order of arguments and relations of the full relationship tensor in its rows and columns. Because the cropping strategy looks at the sub-graph induced by the selected vertices, it maintains the alignment of contextual arguments~(graph's vertices) and relations~(graph's edges). 
% {Because the cropped relationship tensors are resampled in training iteration, this can be viewed as a form of data augmentation.}
The cropped relationship tensors are resampled in each training iteration, which can be viewed as a form of data augmentation.

\subsection{Bidirectional Learning Mechanism}
We have developed a bidirectional learning mechanism to predict the labels of a relationship between its head and tail. 
%encode contextual arguments and their \jesse{ both head and tail relation information.} %two kinds of relationships~(head and tail relation information).
%\jesse{are the contextual arguments and the the head/tail separate? or do you mean contextual head and tail argument info? }
Head and tail relationship information is captured by applying a multi-scale (1D) convolution on the relationship tensor both horizontally and vertically, respectively. 
% Specifically, when we select the $i_{th}$ argument, the head branch utilizes information from all other arguments pointing to the $i_{th}$ argument to facilitate head detection. 
% Similarly, the tail branch can aid in predicting the tail detection. 
% We develop a bidirectional learning mechanism, which consist of a  head convolution and a tail convolution, to extract features from different directions.
% When referring to the argument structure matrix, it is important to note that the direction in which the matrix is oriented can affect its representation. 
% Specifically, when the convolution is applied  vertically versus horizontally on the matrix, the resulting meaning is differ.

\begin{figure}[!ht]
\centering
\includegraphics[width=0.5\linewidth]{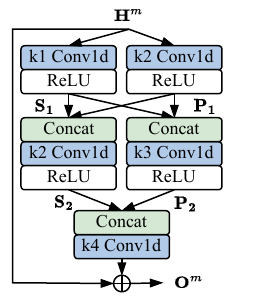}
\caption{The structure of the multi-scale residual module~(MSRM). k1, k2, k3 and k4 represent the kernel size of the respective 1D convolution layer. }
\captionsetup{justification=centering}
\label{fig:msrm}
\end{figure}

Because the relationship tensor can take into account both short- and long-distance relationships between the argument sentences in a discourse, we leverage a multi-scale residual module~(MSRM). 
% The reason why we choose the MSRM to capture varied relations is because its two bypass structure can provide more convolution combinations with limited convolution layers. 
Prior work~\cite{li2020icd} reveals that a multi-filter convolutional layer~(similar to an Inception block) can capture varied relationships.
However, \cite{li2018multi} stated that the Inception architecture leads to the underutilization of local features.
Therefore, we choose the MSRM~\cite{li2018multi}  as our base module 
% and 
% modify it
%based on the first idea above to enhance the ability of the MSRM to capture flexible argument relationships 
and {use different kernel sizes for the convolutional filters  to capture the short- and long distance relationships.}
% The reason why the modified MSRM can capture varied relations is because its varied convolution kernel 
Because of the asymmetry of the relationships, we apply the MSRM both horizontally and vertically on the relationship tensor to generate predictions while taking into account the contextual %neighboring 
argument structure of the discourse. 
%and two kinds of relation information. 
Figure~\ref{fig:msrm} illustrates the structure of the MSRM.

{We illustrate here how to capture the contextual arguments and head relationship information using the horizontal branch during trainng with a cropped tensor.} %is captured. 
%take the horizontal branch as an example to show how the contextual arguments and head relation information is captured. 
Firstly, we obtain the representations of the first row of the relationship tensor,~$\mathbf{h}_1 = \mathbf{H}^{'}[1,:,:] \in \mathbb{R}^{1 \times m \times d}$ and pass them into the MSRM.
We denote the 1D convolutional layer with kernel size $k_x$ as $\mathcal{F}_{k_x}(\cdot)$.
We omit the bias item in the equations and the output~$\mathbf{o}_1 \in \mathbb{R}^{1 \times m \times d}$ is:
\begin{align}
\label{eq:feature_abstract}
\mathbf{S}_1 &= \operatorname{ReLU}(\mathcal{F}_{k_1}(\mathbf{h_1})), \\
\mathbf{P}_1 &= \operatorname{ReLU}(\mathcal{F}_{k_2}(\mathbf{h_1})), \\
\mathbf{S}_2 &= \operatorname{ReLU}(\mathcal{F}_{k_2}(\operatorname{Concat}(\mathbf{S}_1, \mathbf{P}_1))) ,\\
\mathbf{P}_2 &= \operatorname{ReLU}(\mathcal{F}_{k_3}(\operatorname{Concat}(\mathbf{S}_1, \mathbf{P}_1))),\\
\mathbf{o}_1 &= \mathcal{F}_{k_4}(\operatorname{Concat}(\mathbf{S}_2, \mathbf{P}_2)) + \mathbf{h_1},
\end{align}
where $\operatorname{ReLU}(\cdot)$ represents the ReLU activation function and $\operatorname{Concat}(\cdot)$ is the feature concatenation operation. 
%Secondly, \jesse{this is unclear - do you repeat on each row m times or apply it to once to every row yielding m repetitions? Again, it is important that things are unambiguous}
We repeat the above calculations for $m$ times and we concatenate all output features to get the resulting tensor of the head convolution~$\mathbf{O}^{m}_{h} = \operatorname{Concat}(o_1, \cdots, o_m) \in \mathbb{R}^{m \times m \times d}$.
{Thirdly, a linear classifier layer consisting of flattened logits are applied to produce the predictions. On the  horizontal branch, this is $\mathbf{\hat{y}}^{h} \in \mathbb{R}^{l}$ where $l$ represents the label space dimension.} 
%Thirdly, with a linear classifier layer, flattened logits for the horizontal branch is $\mathbf{\hat{y}}^{h} \in \mathbb{R}^{m^2 \times l}$ where $l$ represents the label space dimension. 
{The vertical branch uses the same operations as above but the logits are denoted as $\mathbf{\hat{y}}^{t} \in \mathbb{R}^{l}$.\footnote{Note that during testing (inference stage) in the above computations $m$ is replaced by $n$ (number of paired arguments of the full text) and $\mathbf{H}^{'}$ by $\mathbf{H}$.}}
%To compute the logits for the vertical branch, we use the same operations as above. These logits are denoted as $\mathbf{\hat{y}}^{t} \in \mathbb{R}^{m^2 \times l}$.\footnote{Note that in the above computations $m$ is replaced by $n$ (number of paired arguments of the full text) during testing (inference stage).} 

The cross-entropy losses of the horizontal and vertical branches are calculated as follows:
\begin{align}
\label{eq:M_update_theta}
\mathcal{L}_{h} = - \frac{1}{m^2} \sum_{i=1}^{m^2}\sum_{j=1}^{l} {y}_{i,j} \log(\hat{y}^{h}_{i,j}),  \\
\mathcal{L}_{t d} = - \frac{1}{m^2} \sum_{i=1}^{m^2}\sum_{j=1}^{l} {y}_{i,j} \log(\hat{y}^{t}_{i,j}), 
\end{align}
where $\mathbf{y} \in \mathbb{R}^{l}$ are one-hot encoded vectors representing the ground-truth labels.
We adopt joint training for the two branch losses and the joint training loss is defined as:
\begin{align}
\mathcal{L} = \lambda_{h} \mathcal{L}_{h} + \lambda_{t} \mathcal{L}_{t},
\end{align}
where $\lambda_{h}$ and $\lambda_{t}$ are scaling factors for the horizontal and vertical branches, respectively.

\subsection{Label Fusion}
%\jesse{Is this only at prediction time or is also used during training?}
During testing,
a confidence voter fuses the label predictions from the horizontal and vertical branches. 
% As different branches are good at different prediction modes, i.e., head and tail predictions, w
% We designed confidence voter to merge two prediction matrices.
Inspired by \citet{vyas2018out} and \citet{weng2023boosting}, we measure the confidence scores of logits by the difference between top-$1$ and top-$2$ probabilities. 
Assume that we have logits $\widetilde{\mathbf{y}^{h}} \in \mathbb{R}^{l}$ and $\widetilde{\mathbf{y}^{t}} \in \mathbb{R}^{l}$,
%, and we applied the label fusion mechanism over them.
%Firstly, we use Softmax function to normalize logits and 
the difference between the top-$1$ and top-$2$ probabilities for $\widetilde{\mathbf{y}^{h}}$ is $\widetilde{\mathbf{c}^{h}}$: 
\begin{align}
\widetilde{\mathbf{s}^{h}} &= \operatorname{Softmax}(\widetilde{\mathbf{y}^{h}}), \\
\widetilde{\mathbf{c}^{h}} &= \operatorname{Topk}(\widetilde{\mathbf{s}^{h}}, k_1) - \operatorname{Topk}(\widetilde{\mathbf{s}^{h}}, k_2),
\end{align}
% \jesse{Is this really what you do? by your definition top topk(...,1) returns 1 element and topk(...,2) returns two element. I think you mean it returns the $k^{th}$ largest element. This would be more standard.}
where $\operatorname{Topk}(\cdot , k)$ returns the $k-$th largest elements from the given input and $\operatorname{Softmax}(\cdot)$ is the softmax function.\footnote{Results with other normalization functions (e.g., L1, L2, min-max) did not improve results.}  
Secondly, we get the confidence score for the vertical branch~$\widetilde{\mathbf{c}^{t}}$ by following the above computations.
Thridly, we 
%\jesse{you have switched tenses here} 
applied the argmax operation to get the predictions of two branches, $\overline{\mathbf{y}^{h}} \in \mathbb{R}^{n \times n}$ and $\overline{\mathbf{y}^{t}} \in \mathbb{R}^{n \times n}$.
We merge two matrices into $\overline{\mathbf{y}} \in \mathbb{R}^{n \times n}$ based on confidence scores $\widetilde{\mathbf{c}^{h}}$ and $\widetilde{\mathbf{c}^{t}}$.
The merging process for the $\overline{{y}}_{i,j}$ is:
\begin{align}
    \overline{{y}}_{i,j} = 
\begin{cases}
    \overline{{y}^{h}}_{i,j},& \text{if } \widetilde{{s}^{h}}_{i,j} \geq \widetilde{{s}^{t}}_{i,j}\\
    \overline{{y}^{t}}_{i,j}, & \text{otherwise}
\end{cases}
\end{align}

\begin{table*}[htbp]
\setlength\tabcolsep{2pt} % let LaTeX figure out amount of inter-column whitespace
\begin{center}
\begin{tabular*}{\linewidth}{@{\extracolsep{\fill}} l r|r|r|r|r|r|r|r|r|r|r|r }
\toprule
\multicolumn{1}{l}{\multirow{2}{*}{Models}} & 
\multicolumn{4}{c}{\textbf{Neoplasm}} & \multicolumn{4}{c}{\textbf{Glaucoma}} &\multicolumn{4}{c}{\textbf{Mixed}}\\
& \multicolumn{1}{l}{F1} & \multicolumn{1}{l}{S-F1} & \multicolumn{1}{l}{A-F1} & \multicolumn{1}{l|}{U-F1}  & \multicolumn{1}{l}{F1} & \multicolumn{1}{l}{S-F1} & \multicolumn{1}{l}{A-F1} & \multicolumn{1}{l|}{U-F1} & \multicolumn{1}{l}{F1} & \multicolumn{1}{l}{S-F1} & \multicolumn{1}{l}{A-F1} & \multicolumn{1}{l}{U-F1}     \\ 
\midrule
% \textbf{DMON-SA}& 31.88 & 0.00  &  0.00 &  95.66  & 32.13 & 1.24  &  0.00 &  95.12 & 32.12 &  0.67 & 0.00  &  95.73 \\
% \textbf{TreeAM}& 37.00 & -  &  - &  -  &  45.00 &  - & -  &  - & 40.00 & -  &  - &  - \\
% \textbf{ResidualAM}& 43.00 & -  &  - &  -  &  39.00 &  - & -  &  - & 45.00 & -  &  - &  - \\
%\textbf{RESARG} & 63.16 & -  &  - &  -  & 61.86 &  - & -  &  - & 68.35 & -  &  - &  - \\
\textbf{AMPERE++} & 63.73 & -  &  - &  -  & - &  - & -  &  - & - & -  &  - &  - \\
\textbf{Roberta}& 67.00 & -  &  - &  -  &  66.00 &  - & -  &  - & 69.00 & -  &  - &  - \\
\textbf{AMCT-Sci}& 68.16 & 59.99  & 49.12  &  95.45  & 62.28  & 64.71  &  24.78 &  95.24 & 69.43 &  55.31 &  58.00 &  94.76 \\
\textbf{TransforMED}& 69.96 & 58.72  &  55.65 &  95.51  &  69.72 &  65.32 & 47.00  &  96.88 & 71.82 & 57.14  &  \textbf{63.41} &  94.90 \\
\textbf{RESATTARG}& 70.92 & 52.77  &  \textbf{65.38} &  94.54  &  68.40 &  54.73 & \textbf{56.00}  &  94.36 & 67.66 & 49.62  &  59.09 &  94.21 \\
\hline
% \textbf{DMON-2D}& 74.06 & 68.11  &  57.21 &  97.21  &  70.93 &  72.07 & 44.14  & 97.11 &72.01 & 66.37 & 52.96  &  96.82 \\
\textbf{DMON}& \makecell[c]{\textbf{76.30}\\ $\pm$0.71} &  \makecell[c]{\textbf{68.25}\\ $\pm$0.98} &  \makecell[c]{64.13\\ $\pm$0.47} &  \makecell[c]{\textbf{97.33}\\ $\pm$0.08} &  \makecell[c]{\textbf{74.16} \\ $\pm$1.10} &  \makecell[c]{\textbf{73.16} \\ $\pm$1.17} &  \makecell[c]{53.41 \\ $\pm$0.63} &  \makecell[c]{\textbf{97.27}\\ $\pm$0.06} &  \makecell[c]{\textbf{74.07}\\ $\pm$1.11} &  \makecell[c]{\textbf{68.35}\\ $\pm$1.42} &  \makecell[c]{54.05\\ $\pm$0.74} &  \makecell[c]{\textbf{97.08}\\ $\pm$0.09} \\
% for the evidence search by bm25 
% 29.41 & 10.56 & 20.75 & 25.98 & 8.396 --> case
% 22.14 & 7.34 & 16.62 & 2.644
\bottomrule
\end{tabular*}
\end{center}
% \captionsetup{justification=centering}
\caption{Experimental results of argument structure learning in terms of macro-F1 scores on three AbstRCT medical datasets.
S-F1, A-F1, U-F1 and F1 refer to the average macro-F1 score of the support relation, of the attack relation, of no-relation, and of their average, respectively. For DMON the mean results over 5 runs with variance are shown.
}
\label{table:abs_results}
\end{table*}

\section{Experiments}
\label{sec:exp}
%\jesse{Maybe give teh high level goals or objectives of the experiments?}
The goal is to evaluate the Macro F1 score~(it returns objective results on imbalanced ASL datasets) of the detection and correct classification of argumentative relationships in a discourse, to compare the results with the results of state-of-the-art baselines, and to assess the influence of the proposed components of the DMON model. 
\subsection{Datasets}
We conduct experiments in the medical, legal and scientific domains, represented by the AbstRCT~\cite{mayer2020transformer}, the Cornell eRulemaking Corpus~(CDCP)~\cite{niculae2017argument}, and the SciDTB~\cite{yang2018scidtb} datasets.

\noindent \textbf{AbstRCT:}  The AbstRCT corpus consists 659 medical documents about the treatment for specific disease~(neoplasm, glaucoma, hypertension, hepatitis-b, diabetes).
Following \citet{mayer2020transformer}, the corpus is divided into three datasets based on disease category: neoplasm, glaucoma, and mixed.
The neoplasm~(neo) dataset contains 350 documents for training, 50 for validation and 100 notes for testing. The neoplasm train set is used as train set for the glaucoma~(gla) and mixed~(mix) dataset, which each have 100 instances for testing. 
%The glaucoma dataset has 100 samples for testing.
%The mixed dataset consists of medical documents for five different disease and the dataset also has 100 instances for testing. 
The labels of the relationships of the three AbstRCT datasets are `attack', `support', and `un-related'. 

\noindent \textbf{SciDTB:} The SciDTB~(SCI) dataset includes 1049 scientific abstracts collected from the ACL Anthology.
It consists of 743 examples for training, 154 samples for validation and 152 for testing. 
The dataset has more fine-grained discourse relationship categories while the number of labels is 27.
% (see Table \ref{table:scidtb_space}).
%The label set is given in the appendix~\ref{sec:res_scidtb}.

\noindent \textbf{CDCP:} The CDCP dataset contains 731 user comments about consumer debt collection practices from an eRulemaking website and include 581 examples for training and 150 for testing. 
Labels of the CDCP dataset are `related' and `un-related'. 

% \noindent \textbf{SciDTB:} The SciDTB dataset includes 1049 scientific abstracts collected from the ACL Anthology.
% It consists of 743 examples for training, 154 samples for validation and 152 for testing. 
% The dataset has more fine-grained discourse relationship categories while the number of the dataset's labels is 27.
% A full label set is on the appendix~\ref{sec:res_scidtb}.

\subsection{Experimental Set-up}

\noindent \textbf{Model Settings:} 
The maximum sequence length is $128$. 
%Following \cite{li2018multi} and promoting diversity in kernel sizes, the combination of kernel sizes for the multi-scale convolution module is set as $\{k_1, k_2, k_3, k_4, k_5\} = \{7, 5, 5, 3, 1\}$.
The combination of kernel sizes for the multi-scale convolution module is set to $\{k_1, k_2, k_3, k_4, k_5\} = \{7, 5, 5, 3, 1\}$.
The window size for the cropping is $13$. $\lambda_{h}$ and $\lambda_{t}$ are set to 0.5.
$k_1$ and $k_2$ is set as 1 and 2 in the label fusion part.

\noindent \textbf{Training Details:} 
We fine-tune the BioLinkBERT~\cite{yasunaga2022linkbert} for the AbstRCT dataset and LinkBERT~\cite{yasunaga2022linkbert} for the SciDTB and CDCP dataset.  
For the training, all baseline models and our framework are trained with FP16.
We train all models on a \textit{NVIDIA GeForce RTX 3090} GPU.
% Due to computation limitation, we set batch size as 1 and the gradient accumulation steps as 2.
All models use the AdamW optimizer~\cite{loshchilov2017decoupled} with a learning scheduler initialized at $2e^{-5}$ and linearly decreased to 0.

\begin{table}[htbp]
\setlength\tabcolsep{3pt} % let LaTeX figure out amount of inter-column whitespace
\begin{center}
\begin{tabular}{l |cccc}
\toprule
\multirow{1}{*}{Models} & Full-F1 & F1 & R-F1 & U-F1\\ 
\midrule
% \textbf{Deep-Res-LG} & - & 29.30 & -  \\
% \textbf{DMON-SA} & 48.35 & 0.26 & 96.44   \\
\textbf{RESARG} & - & 67.60 & 38.99 & 96.20   \\
\textbf{RESATTARG} & - & 73.64 & 50.00& 97.28  \\
\textbf{BERT} & 32.83 & 73.89 & 50.47 & 97.30 \\
\textbf{AMCT-Sci} & 34.58 & 75.89 & 54.61 & 97.17 \\
\textbf{Roberta} & 35.96 & 75.88 & 54.25 & 97.51 \\
% \textbf{LinkBERT} & - & - & - & - \\
\hline
\textbf{DMON} & \makecell[c]{\textbf{48.37}\\ $\pm$0.54} & \makecell[c]{\textbf{87.36}\\ $\pm$0.25} & \makecell[c]{\textbf{77.03}\\ $\pm$0.36} & \makecell[c]{\textbf{98.70}\\ $\pm$0.02}  \\
\bottomrule
\end{tabular}
\end{center}
% \captionsetup{justification=centering}
\caption{Experimental results of argument structure learning in terms of macro-F1 scores on the SciDTB datasets. For DMON the mean results over 5 runs with variance are shown. 
%`F1, R-F1, U-F1' refers to the average macro-F1 score, macro-F1 score of the related relation and no-related relation. 
Full-F1, R-F1, U-F1 and F1 refer to the average macro-F1 score of the full label space, the related relation, of no-relation, and of related and no-relation, respectively. For DMON the mean results over 5 runs with variance are shown. }
\label{table:scidtb_result}
\end{table}

\subsection{Baselines}

% \noindent \textbf{DMON-SA} replaced the bidirectional learning mechanism as a self-attention module to capture both head and tail information.

% \noindent \textbf{TreeAM}~\cite{eger2017neural} regards argument structure learning as a dependency parsing problem and leverage TreeLSTM to extract argument structure from text. 

% \noindent \textbf{ResidualAM}~\cite{galassi2018argumentative} applied a residual neural network to conduct pairwise argument structure learning.

%\noindent \textbf{RESARG}~\cite{galassi2021multi} used a BiLSTM to extract textual feature and then applied a residual  neural network to extract more semantic features for pairwise argument structure learning. 
We consider models that classify the argument relationships given a representation of the pairs of sentences obtained with a pretrained encoder. 
% \noindent \textbf{AMPERE++}~\cite{hua2022efficient} uses a Roberta model to represent sentences where  broader context with window size 20, that is, neighboring sentences are concatenated to the input pair of arguments.  

\noindent \textbf{AMPERE++}~\cite{hua2022efficient} uses a Roberta model to concatenate $20$ neighbouring sentences with an argument and only predicts whether this argument is a head or tail argument. 
The authors did not name this model so we call it AMPERE++.
%designed a context-aware Transformer-based model to conduct pairwise argument structure learning.
%Specifically, they used Roberta encoders to incorporate more context information, i.e., other arguments. 

\noindent \textbf{BERT}~\cite{devlin2018bert} fine-tunes a pretrained BERT model to encode an argument pair and predict its relationship. 

\noindent \textbf{Roberta}~\cite{liu2019roberta} fine-tunes a pretrained Roberta model to represent a pair of sentences and classifies the relationships. 

%sentences conducted pairwise ASL with a pretrained Roberta model~\footnote{\url{https://huggingface.co/roberta-base}}. 

\noindent \textbf{AMCT-Sci}~\cite{stylianou2021transformed} is similar to Roberta, but encodes argument pairs with a domain-specific BERT model.\footnote{\url{https://huggingface.co/allenai/scibert_scivocab_uncased}}

\noindent \textbf{TransforMED}~\cite{stylianou2021transformed} 
% designed an Evidence-Based Medicine~(EBM) model and use it to inject additional medical knowledge into the pairwise ASL model. 
integrates a medical knowledge system to extract medical entities from arguments.
The authors inject medical knowledge into their model by concatenating features of arguments and medical entities. 
% The TransforMED is the current state-of-the-art model on three AbsRCT medical datasets. 

\noindent We also consider models that use attention mechanisms to model relationships between arguments. 

\noindent \textbf{RESARG}~\cite{galassi2021multi} used a BiLSTM to extract textual feature and then applied a residual neural network to deal with the ASL task. 

\noindent \textbf{RESATTARG}~\cite{galassi2021multi} extended RESARG with a coarse-grained parallel co-attention mechanism to predict argumentative relations. 

\noindent \textbf{TSP-PLBA}~\cite{morio2020towards} consists of task-specific parameterization~(TSP) and proposition-level biaffine attention~(PLBA) to capture argument structure from documents. 
TSP encodes the arguments while PLBA predicts argument relations by using a biaffine scoring function. 
% referred to the coarse-grained parallel co-attention~\cite{galassi2020attention} and designed a attention module.
% They 
% extended their previous work~(RESARG) with the designed attention module and ensemble learning and achieved better performance. 

\noindent We then consider models that train a transition-based dependency parser.
% \noindent \textbf{St-SVM-strict}~\cite{niculae2017argument} propose a factor graph model to extract argument structure from text. 

\noindent \textbf{BERT-Trans}~\cite{bao2021neural} 
% proposed a neural transition-based model to construct argument graph.
% ELMo-Trans leverage ELMo to obtain hidden representation and use the transition system to extract argument structure. 
leverage the BERT language model to obtain representation and propose a neural transition-based model to generate a sequence of actions~(shift, delete-delay, delete, right-arc, right-arc-delay, and left-arc) that build an argument structure~(predicted nodes and relations). 

% \noindent \textbf{BERT-Trans}~\cite{bao2021neural} is similar to ELMo-Trans but uses a BERT model to encode text. 
% The BERT-Trans is the current state-of-the-art model on the CDCP dataset. 

% \noindent \textbf{DMON-2D} replaced the bidirectional learning mechanism with a 2D multi-scale convolution module to capture both head and tail information.

% \begin{table}[htbp]
% \setlength\tabcolsep{7pt} % let LaTeX figure out amount of inter-column whitespace
% \begin{center}
% \begin{tabular}{l |ccc}
% \toprule
% \multirow{1}{*}{Models} & F1 & R-F1 & U-F1\\ 
% \midrule
% % \textbf{Deep-Res-LG} & - & 29.30 & -  \\
% \textbf{DMON-SA} & 48.35 & 0.26 & 96.44   \\
% \textbf{St-SVM-strict} & - & 26.70 & -   \\
% \textbf{TSP-PLBA} & - & 34.00 & - \\
% \textbf{AMPERE++} & 63.10 & - & -  \\
% %\textbf{RESARG} & 63.40 & 28.60 & 98.20  \\
% \textbf{RESATTARG} & 64.40 & 30.60 & 98.30  \\
% \textbf{ELMo-Trans} & 67.10 & 35.60 & \textbf{98.60}  \\
% \textbf{DMON-2D} &  67.25 & 36.36 &  98.34  \\
% \textbf{BERT-Trans} & 67.80 & 37.30 & 98.30    \\
% \hline
% \textbf{DMON} & \makecell[c]{\textbf{68.14}\\ $\pm$0.45} & \makecell[c]{\textbf{38.26}\\ $\pm$0.74} & \makecell[c]{{98.37}\\ $\pm$0.06}  \\
% \bottomrule
% \end{tabular}
% \end{center}
% % \captionsetup{justification=centering}
% \caption{Experimental results of argument structure learning in terms of macro-F1 scores on the CDCP datasets. `F1, R-F1, U-F1' refers to the average macro-F1 score, macro-F1 score of the related relation and no-related relation. 
% }
% \label{table:cdcp_result}
% \end{table}

\begin{table}[htbp]
\setlength\tabcolsep{7pt} % let LaTeX figure out amount of inter-column whitespace
\begin{center}
\begin{tabular}{l |ccc}
\toprule
\multirow{1}{*}{Models} & F1 & R-F1 & U-F1\\ 
\midrule
% \textbf{Deep-Res-LG} & - & 29.30 & -  \\
% \textbf{DMON-SA} & 48.35 & 0.26 & 96.44   \\
% \textbf{St-SVM-strict} & - & 26.70 & -   \\
\textbf{TSP-PLBA} & - & 34.00 & - \\
\textbf{AMPERE++} & 63.10 & - & -  \\
%\textbf{RESARG} & 63.40 & 28.60 & 98.20  \\
\textbf{RESATTARG} & 64.40 & 30.60 & 98.30  \\
% \textbf{ELMo-Trans} & 67.10 & 35.60 & \textbf{98.60}  \\
% \textbf{DMON-2D} &  67.25 & 36.36 &  98.34  \\
\textbf{BERT-Trans} & 67.80 & 37.30 & 98.30    \\
\hline
\textbf{DMON} & \makecell[c]{\textbf{68.14}\\ $\pm$0.45} & \makecell[c]{\textbf{38.26}\\ $\pm$0.74} & \makecell[c]{\textbf{98.37}\\ $\pm$0.06}  \\
\bottomrule
\end{tabular}
\end{center}
% \captionsetup{justification=centering}
\caption{Experimental results of argument structure learning in terms of macro-F1 scores on the CDCP datasets. For DMON the mean results over 5 runs with variance are shown. 
%`F1, R-F1, U-F1' refers to the average macro-F1 score, macro-F1 score of the related relation and no-related relation. 
R-F1, U-F1 and F1 refer to the average macro-F1 score of the related relation, of no-relation, and of their average, respectively.}
\label{table:cdcp_result}
\end{table}

% \begin{table}[htbp]
% \setlength\tabcolsep{7pt} % let LaTeX figure out amount of inter-column whitespace
% \begin{center}
% \begin{tabular}{l |ccc}
% \toprule
% \multirow{1}{*}{Models} & F1 & R-F1 & U-F1\\ 
% \midrule
% % \textbf{Deep-Res-LG} & - & 29.30 & -  \\
% % \textbf{DMON-SA} & 48.35 & 0.26 & 96.44   \\
% % \textbf{St-SVM-strict} & - & 26.70 & -   \\
% \textbf{TSP-PLBA} & - & 34.00 & - \\
% \textbf{AMPERE++} & 63.10 & - & -  \\
% %\textbf{RESARG} & 63.40 & 28.60 & 98.20  \\
% \textbf{RESATTARG} & 64.40 & 30.60 & 98.30  \\
% \textbf{ELMo-Trans} & 67.10 & 35.60 & \textbf{98.60}  \\
% % \textbf{DMON-2D} &  67.25 & 36.36 &  98.34  \\
% \textbf{BERT-Trans} & 67.80 & 37.30 & 98.30    \\
% \hline
% \textbf{DMON} & \makecell[c]{\textbf{68.14}\\ $\pm$0.45} & \makecell[c]{\textbf{38.26}\\ $\pm$0.74} & \makecell[c]{{98.37}\\ $\pm$0.06}  \\
% \bottomrule
% \end{tabular}
% \end{center}
% % \captionsetup{justification=centering}
% \caption{Experimental results of argument structure learning in terms of macro-F1 scores on the CDCP datasets. For DMON the mean results over 5 runs with variance are shown. 
% %`F1, R-F1, U-F1' refers to the average macro-F1 score, macro-F1 score of the related relation and no-related relation. 
% R-F1, U-F1 and F1 refer to the average macro-F1 score of the related relation, of no-relation, and of their average, respectively.}
% \label{table:cdcp_result}
% \end{table}

\subsection{Results and Discussion}
%We utilize macro-F1 score to evaluate all models. 
% Table~\ref{table:abs_results} shows the experimental results of all models on three absRCT datasets, i.e., Neoplasm, Glaucoma, and Mixed, and Table~\ref{table:cdcp_result} is for the CDCP datasets. 
We run each model five times with different seeds and report the average macro-F1 scores and their variance obtained on the three absRCT datasets the CDCP dataset and the SciDTB dataset.
% \textbf{1)} 

\noindent \textbf{AbsRCT:} Table~\ref{table:abs_results} shows that the DMON outperforms all baselines on all average F1 scores. 
Compared with the state-of-the-art model TransforMED, our model improves the average macro-F1 scores by $6.34$, $5.76$, and $2.25$ percentage points on the Neoplasm, Glaucoma, and Mixed datasets, respectively.
Even though TransforMED explicitly injects external medical knowledge, our approach still performs better.
% We observe that the A-F1 score does not achieve the optimal value, and the reason behind this is the presence of a label imbalance issue in the AbstRCT dataset. 
% Concretely, there are fewer samples annotated as attacks compared to the support samples.
% \textbf{2)} 

\noindent \textbf{SciDTB:} Table~\ref{table:scidtb_result} shows that the DMON outperforms baseline models by a large margin when evaluated on the SciDTB dataset. 
Compared to the Roberta model, our model improves the average macro-F1 scores by $12.41$, $11.48$, and $22.78$ percentage points on Full-F1, F1, and R-F1 scores. 
%The full label space is shown in Table~\ref{table:scidtb_space}.
% Results for the individual labels in terms of macro-F1 scores are in appendix~\ref{sec:res_scidtb}.

% \begin{table}[htbp]
%     \centering
%     \begin{tabularx}{\linewidth}{|X|}
%         \hline
%         \textbf{Labels}    \\ \hline
%         'ROOT', 'enablement', 'elab-aspect', 'cause', 'elab-addition', 'comparison', 'manner-means', 'evaluation', 'bg-compare', 'elab-example', 'joint', 'bg-goal', 'contrast', 'same-unit', 'progression', 'attribution', 'result', 'elab-enum-member', 'temporal', 'bg-general', 'condition', 'elab-process-step', 'exp-evidence', 'elab-definition', 'summary', 'exp-reason'         \\ \hline
%     \end{tabularx}
%     \caption{The full label space for the SciDTB dataset.}
%     \label{table:scidtb_space}
% \end{table}

\noindent \textbf{CDCP:} Table~\ref{table:cdcp_result} shows that DMON also achieves the best performance in terms of average macro-F1 score on the CDCP dataset. 
Compared to the BERT-Trans model 
%does not report the variance of the results in the mean and standard format, for fair comparison we pick the best score, 
our method improves the averaged macro-F1 and R-F1 by $0.79$ and $1.7$ percentage points, respectively.
% BERT-Trans model developed a complicated transition systems with six types of actions.
% However, our model is simple, but it still can achieve better performance compared with the SOTA model.  
Compared with the BERT-Trans, our model is simple~(can be applied to other pairwise classification models) and can achieve better performance. 
Additionally, BERT is the transformer encoder but the proposed neural transition-based model is to generate a sequence of actions (shift, delete-delay, delete, right-arc, right-arc-delay, and left-arc) that build an argument structure (predicted nodes and relations). 
To train this transition system, they need to convert argument structure learning data into the transition-based structure data. 
This preprocessing adds complexity.
% As the BERT-Trans model $1.7$ $0.79$

\noindent \textbf{Discussion:} We observe that the performance gains of DMON oompared to state-of-the-art baselines are different when analyzing the results obtained on the three domains.
%has achieved different performance improvements on 
%three domain datasets.
For instance, when comparing with baseline RESATTARG, DMON improves the macro-F1 scores by 13.92, 5.38, 3.74 percentage points on SciDTB, Neoplasm, and CDCP datasets, respectively. The AbstRCT (of which Neoplasm is a part) and CDCP are imbalanced and have a high ratio of sentence pairs that exhibit no argumentative relationship ("unrelated" relationship). 
This could be a reason why DMON has somewhat lower performance gains. 
On the other hand, AbstRCT and CDCP have few relationship types and it might be that the baselines already deal with these in a satisfactory way, while they have more difficulties with the 27 relationship types of SciDTB. 
For this more difficult case of argumentative structure learning, DMON has the highest gains in performamce and improves the average macro-F1 scores by $12.41$, $11.48$, and $22.78$ percentage points on all 27 relationship type, on the "unrelated" class and on the other 26 relationships, respectively. 

% We explore the efficiency of DMON by measuring its training speed~(sec). 
% RESATTARG, the state-of-the-art model on the AbstRCT dataset takes 1425 sec to finish training while DMON trains within 1240 sec.
% As no implementation of BERT-Trans (i.e., the state-of-the-art model on the CDCP dataset) is publicly available, we compare the training speed of DMON with the one of RESATTARG. 
% DMON is almost three times faster than RESATTARG~(2856 sec). 
% For the SciDTB dataset, the training speed of DMON is 2610 sec and the Roberta model takes 4410 sec. 

\noindent \textbf{Large Language Models:}
As large language models capture the attention of the NLP communities, we conduct a comparison of LLMs with the proposed algorithm.
We pick the GPT 3.5 turbo~(gpt-3.5-turbo-0613), a widely used LLM, and validate it on the ASL task. 
We use the in-context prompt learning~(ICL)~\cite{NEURIPS2020_1457c0d6} to validate the ASL task. 
We found the zero-shot ICL is much worse than the few-shot ICL, so we reported the results of the GPT 3.5 turbo by using the few-shot ICL method. 
Experimental results show the number of the demonstration samples is 2. 
We report the Macro-F1 scores of the GPT 3.5 on the AbstRCT (gla), AbstRCT (neo), AbstRCT (mix), CDCP and SciDTB datasets, which are 12.03, 13.84, 11.19, 11.89, and 17.24, respectively. 
Results of the GPT 3.5 turbo are much lower than the fine-tuned models, i.e., our DMON model, whose scores are 73.16, 76.30, 74.07, 87.36, and 68.14. 
We think the reason why the GPT 3.5 performs badly on the ASL datasets is because LLMs fail to deal with tasks which require complex reasoning ability.

\begin{figure*}[htbp]
\centering
\subfloat[\scriptsize{Neoplasm Dataset}]{\includegraphics[width=0.18\textwidth]{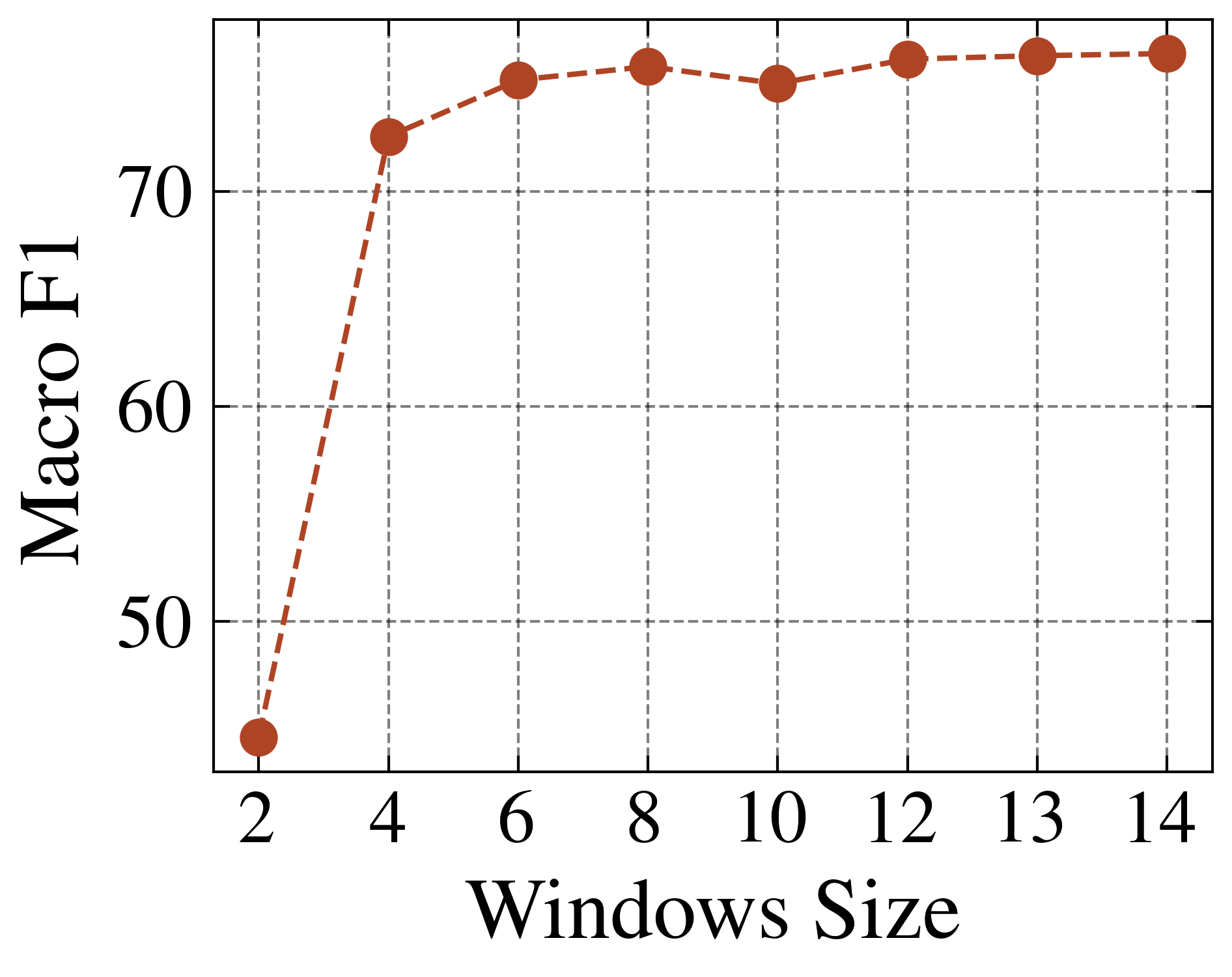}
%\label{fig:compare_entr_EMIN_MHOP_1}
}
\hfil
\subfloat[\scriptsize{Glaucoma Dataset}]{\includegraphics[width=0.18\textwidth]{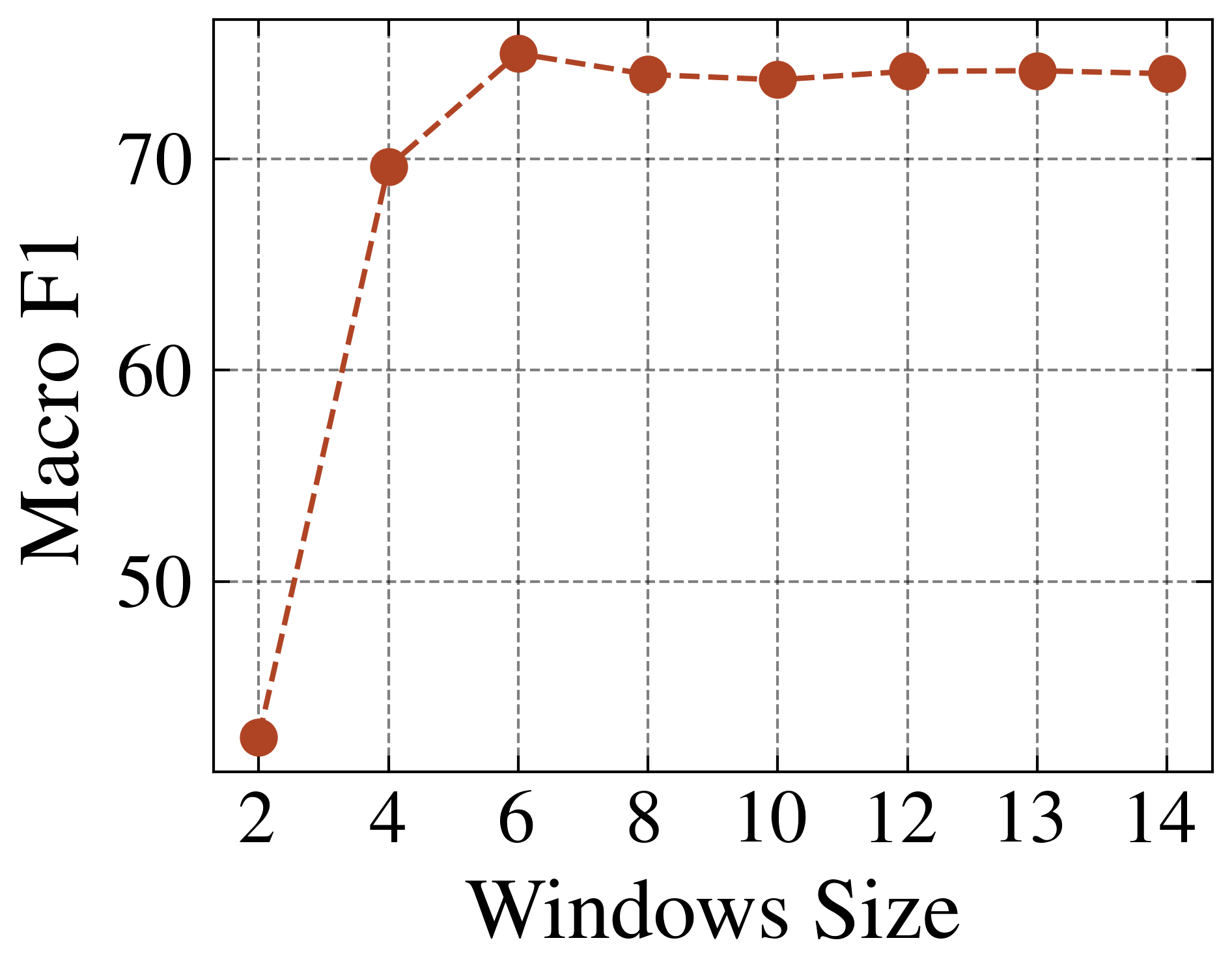}
%\label{fig:compare_entr_EMIN_MHOP_2}
}
\hfil
\subfloat[\scriptsize{Mixed Dataset}]{\includegraphics[width=0.18\textwidth]{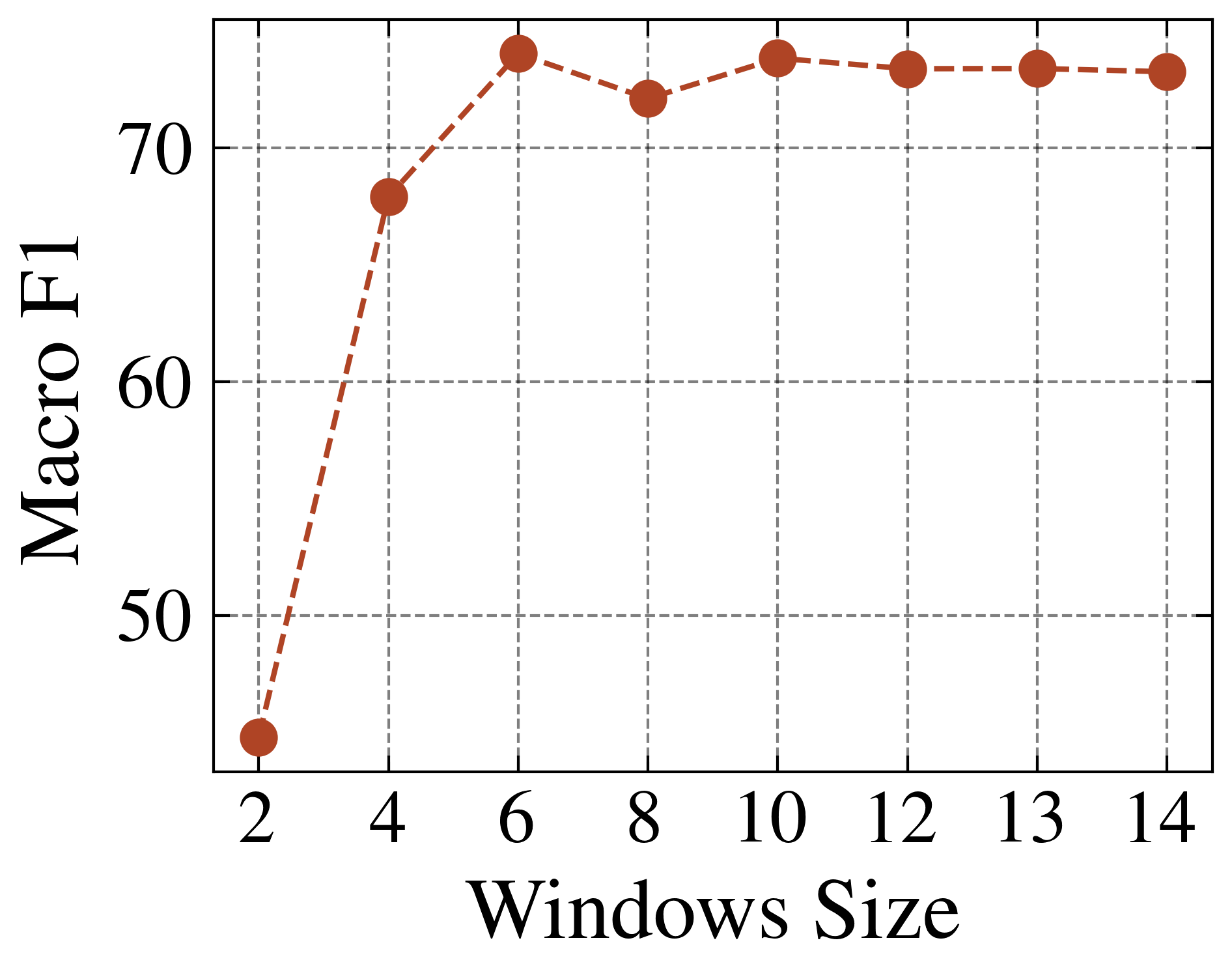}
%\label{fig:compare_entr_EMIN_MHOP_2}
}
\hfil
\subfloat[\scriptsize{CDCP Dataset}]{\includegraphics[width=0.18\textwidth]{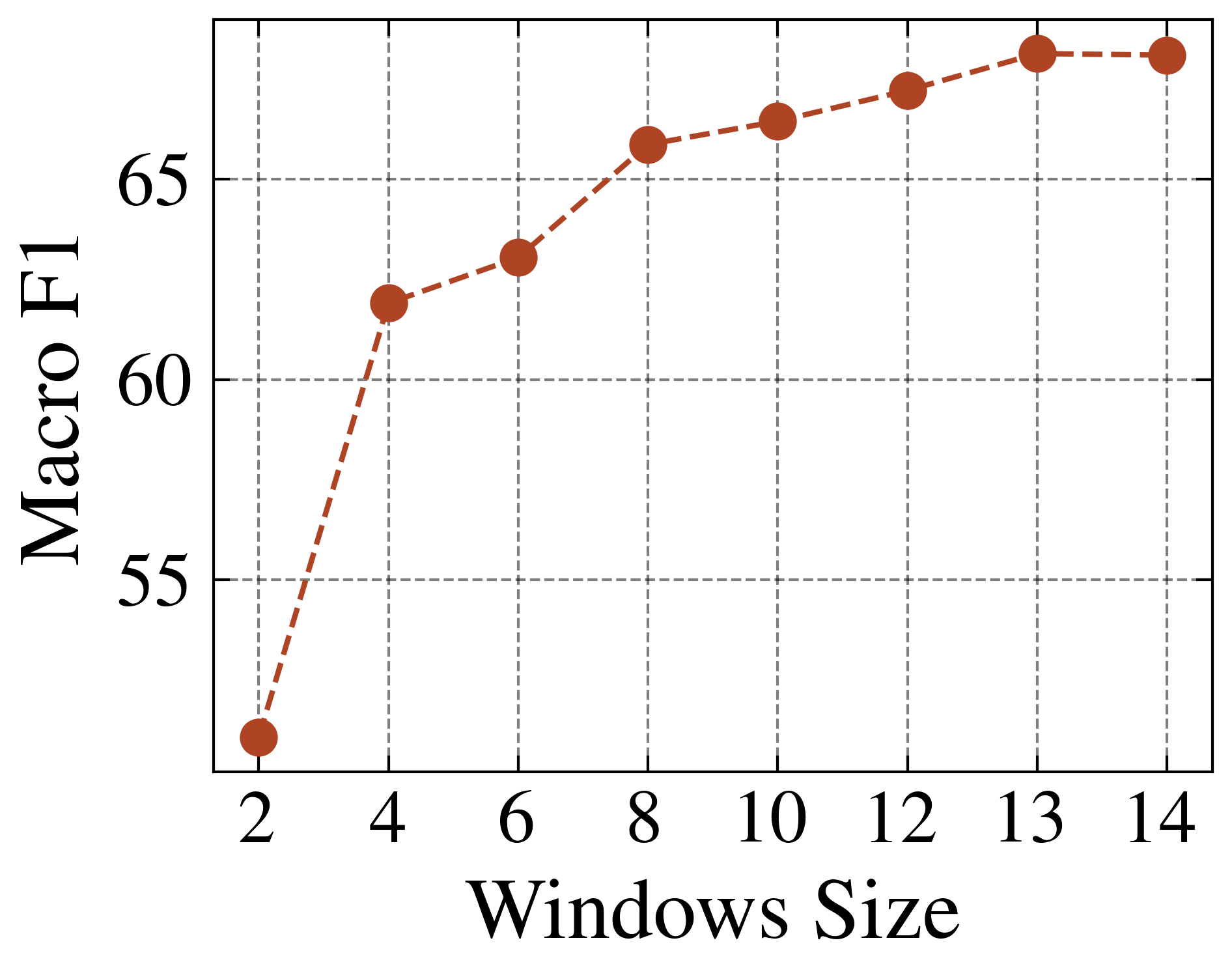}
%\label{fig:compare_entr_EMIN_MHOP_2}
}
\hfil
\subfloat[\scriptsize{SciDTB Dataset}]{\includegraphics[width=0.18\textwidth]{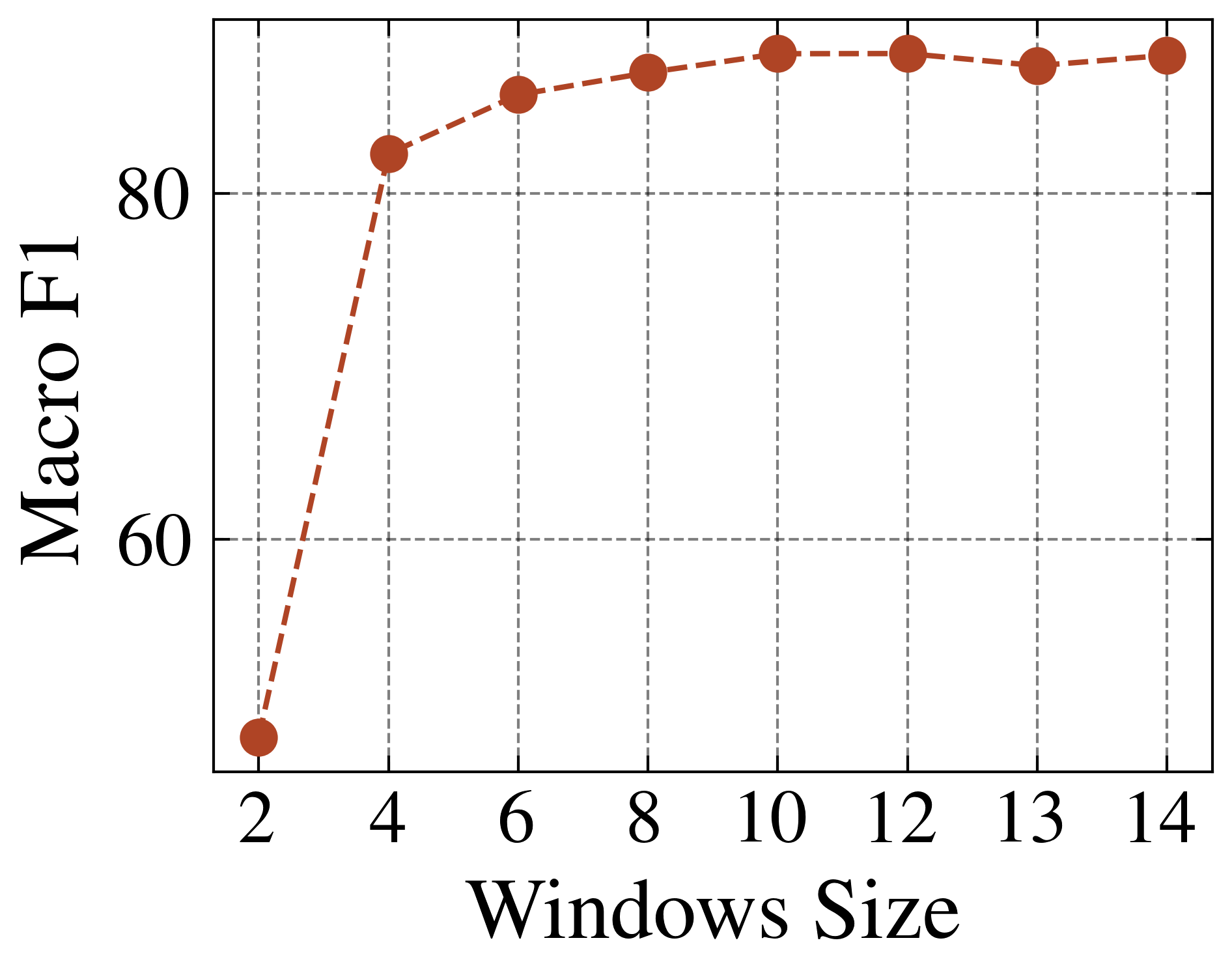}
%\label{fig:compare_entr_EMIN_MHOP_2}
}
\caption{Macro-F1 scores when changing window size of contextual arguments. }
\label{fig:win_size}
\end{figure*}

% random: --> bad performance --> table

\subsection{Ablation Study and Analysis of Results}
We conduct ablation experiments to study DMON's components and analyze them. 
% , i.e., bidirectional learning mechanism, multi-scale %convolution, confidence voter and alignment cropping strategy.

\subsubsection{Bidirectional Learning Mechanism and Confidence Voter}
Table~\ref{table:ablation_results} shows results when ablating the bidirectional learning mechanism and confidence voter. % are shown in Table~\ref{table:ablation_results}.
% Macro-F1 scores on four datasets are decreased when only using predictions of head or tail branch, but the tail branch achieved better score on the Mix dataset.
The F1 scores obtained on all datasets are largely reduced if we remove the bidirectional learning mechanism and confidence voter. 
The macro-F1 scores on all three datasets~(Neoplasm, Glaucoma, and CDCP) decrease when we just use the predictions of one of the two branches (i.e., head or tail relationships) and use no confidence voter. The same pattern holds when we directly remove one of two branches when training the model.
% It shows the same pattern when we remove head or tail branch, i.e., performance on dataset decreased, but tail branch achieved better macro-F1 score on the Mix dataset.
% \jesse{The following sentence is unclear: } The macro-F1 score is reduced on the Mix dataset when we only reply on predictions from the vertical branch or keep the vertical branch.
% \jesse{This seems to be about both learning and conf but you've already discussed the learning so I'd just discuss the conf part now.} 
The macro-F1 score is reduced on the Mixed dataset when we remove the confidence voter and use tail prediction. 
The macro-F1 shows the same pattern when we only remove the head branch.

\begin{table}[htbp]
\setlength\tabcolsep{2.5pt} % let LaTeX figure out amount of inter-column whitespace
\begin{center}
\begin{tabular}{l |ccccc}
\toprule
\multirow{1}{*}{Model} & Neo & Gla  & Mix & CDCP & SCI\\ 
\midrule
% \textbf{Deep-Res-LG} & - & 29.30 & -  \\
\textbf{DMON} & \textbf{76.30} & \textbf{74.16} & {74.07} & \textbf{68.14} & \textbf{87.36} \\
\hline
\textit{w/o} Voter~(h) & 74.31 & 72.04 & 73.42  & 67.50 & 87.08 \\
\textit{w/o} Voter~(t) & 75.74 & 73.83 & 74.84  & 67.50 & 86.58\\
\textit{w/o} T & 73.74 & 70.10 & 71.40 & 65.39 & 85.75\\
\textit{w/o} H & 75.62 & 71.48 & \textbf{74.88} & 66.55 & 87.08\\
\textit{w/o} H+T & 69.10 & 71.28 & 70.35  & 56.24 & 74.27\\
% \textit{w/o} Group & - & - & -  & \\
\bottomrule
\end{tabular}
\end{center}
% \captionsetup{justification=centering}
\caption{Experimental results of argument structure learning in terms of average macro-F1 scores on the Neoplasm~(Neo), Glaucoma~(Gla), Mixed~(Mix), and CDCP datasets.
\textit{w/o} Voter~(h) or Voter~(t) removes the confidence voter and leverages head or tail prediction, respectively.
\textit{w/o} T or H removes either the head branch or tail branch when training the model.
\textit{w/o} H+T completely removes the bidirectional learning mechanism.
}
\label{table:ablation_results}
\end{table}

\subsubsection{Cropping Strategy}
We analyze several aspects of the cropping strategy.
%with two aspects: data augmentation, information alignment, and neighboring window size. 

\noindent \textbf{Training with Different Cropped Tensors:} Figure~\ref{table:compare_full} compares results of DMON using cropped tensors with the results of using the full relationship tensor as input. 
We observe that DMON using cropped tensors outperforms DMON with the full tensor on the four datasets.
Using the cropping strategy improves the macro-F1 score of our framework by $0.2$, $0.95$, $4.25$, $1.38$, and $0.18$ percentage points on Neoplasm, Glaucoma, Mixed and SciDTB datasets, respectively. The cropped tensors offer more variation in the training data, and its representations contribute to the generalization capabilities of the model.   

\begin{table}[h]
\setlength\tabcolsep{2.5pt} % let LaTeX figure out amount of inter-column whitespace
\begin{center}
\begin{tabular}{l |ccccc}
\toprule
\multirow{1}{*}{Model} & Neo & Gla  & Mix & CDCP & SCI\\ 
\midrule
% \textbf{Deep-Res-LG} & - & 29.30 & -  \\
$\textbf{DMON}_{\text{win=13}}$ & \textbf{76.30} & \textbf{74.16} & \textbf{74.07} & \textbf{68.14} & \textbf{87.36}\\
\hline
$\textbf{DMON}_{full}$ & 76.10 & 73.21 & 69.82  & 66.76 & 87.18\\
% \textit{w/o} Group & - & - & -  & \\
\bottomrule
\end{tabular}
\end{center}
% \captionsetup{justification=centering}
\caption{Results of the cropping strategy with window size of 13 ($\textbf{DMON}_{\text{win=13}}$) and of using the complete relationship tensor ($\textbf{DMON}_{full}$) during training.
%means the input data of the DMON is . 
} 
\label{table:compare_full}
\end{table}

\begin{figure}[htbp]
\centering
\includegraphics[width=0.9\linewidth]{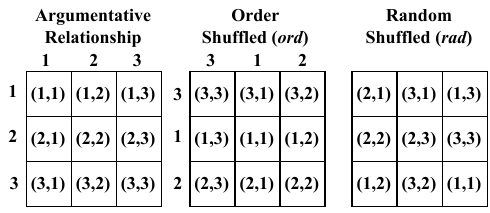}
\caption{An example that illustrates the order shuffle and random shuffle. For simplicity we only show the relationship matrix.}
\captionsetup{justification=centering}
\label{fig:shuffle_demo}
\end{figure}

\noindent \textbf{Information Alignment:} To study the information alignment of the cropping with the original relationship tensor, we develop two shuffle approaches called order shuffle~(\textit{ord}) and 
%symmetry shuffle~(\textit{sym}).
random shuffle~(\textit{rad}).
\textit{ord} scatters the order of the arguments, that is, horizontal and vertical indexes of the cropped relationship tensor are shuffled. 
\textit{rad} randomly chooses pairwise samples to fill the cropped relationship tensor. 
Figure~\ref{fig:shuffle_demo} shows an example illustrating the \textit{ord} and \textit{rad} approaches.
Table~\ref{table:random} shows that the macro-F1 score decreases by using the \textit{rad} method.
Therefore, it is important to keep head and tail relationships aligned, that is, correctly representing the asymmetric relationships between arguments. 
% not breaking the symmetry structure of argumentative relationship matrices. 
Table~\ref{table:random} also reveals that the order of the argumentative sentence pairs in the relationship tensor is important as it implicitly captures the coherence of the discourse.

\begin{table}[htbp]
\setlength\tabcolsep{1.5pt} % let LaTeX figure out amount of inter-column whitespace
\begin{center}
\begin{tabular}{c |c|c|ccccc}
\toprule
\textit{original} & \textit{ord} & \textit{rad} &Neo & Gla  & Mix & CDCP & SCI\\ 
\midrule
% \textbf{Deep-Res-LG} & - & 29.30 & -  \\
 \cmark & \xmark & \xmark & \textbf{76.30} & \textbf{74.16} & \textbf{74.07} & \textbf{68.14} & \textbf{87.36} \\
 \hline
 \xmark & \cmark & \xmark & 75.20 & 68.68 & 72.44  & 61.07 & 75.88 \\
 \xmark & \xmark & \cmark & 71.43 & 74.31 & 72.85  & 65.86 & 86.69  \\
 \xmark & \cmark & \cmark & 59.98 & 55.08 & 60.02  & 50.78 & 66.27 \\
\bottomrule
\end{tabular}
\end{center}
% \captionsetup{justification=centering}
\caption{An experiment for exploring the
information alignment of the cropping strategy.
\textit{original} represents DMON without any shuffling strategies. 
}
\label{table:random}
\end{table}

\noindent \textbf{Contextual Windows Size:} The results demonstrate that 
%As demonstrated by the results, it is beneficial for the ASL task by 
encoding contextual arguments and their relationships is beneficial. 
Figure~\ref{fig:win_size} shows the macro-F1 scores by changing the window size of the cropped relationship tensor for the evaluated datasets. 
The macro-F1 score curve shows an upward trend by encoding more neighboring arguments and relationships, but slightly decreases when considering the full discourse (Table \ref{table:compare_full}).

\section{Conclusion}
\label{sec:conclude}

In this paper, we have proposed a novel framework called the Dual-tower Multi-scale cOnvolution neural Network (DMON) to deal with the ASL task that in a flexible way can learn the argumentative DAG structure taking into account contextual argumentative relationships. A sentence or clause on its own seldom fulfils an argumentative role in a discourse, it is only when paired with another sentence and in the context of other sentences that its argumentative role becomes apparent. In an argumentative DAG structure a sentence can have multiple parents and children, and our model can deal with this flexibility.   
We conduct experiments on four datasets covering the medical, legal and scientific domains, namely abstRCT, CDCP SciDTB and achieve new state-of-the-art performance, when compared to several strong baselines. 
%During training, we represent the argumentative structure with a symmetric cropped relationship matrix to easily encode neighboring arguments and relationships.
%We designed the bidirectional learning mechanism to capture neighboring information for benefiting the ASL task. 
%The symmetry cropping strategy is to deal with data scarcity problem of the ASL datasets. 
Furthermore, we perform ablations and in-depth analyses to prove the effectiveness of each component of our model.
%Our model can be a competitive baseline model not only on the ASL task, but also other NLP tasks including discourse segment ranking and pronoun processing. 
% 
% To bring the broader reception field into the pairwise argument relation classification, we simply use convolution 
% The main reason is because they ignore~\cite{stylianou2021transformed} or fail~\cite{hua2022efficient} to capture argument structure.
% Recently, advanced works 

\section*{Limitations}
The limitation of our paper are reflected as follows:
\textbf{1)} Computation limitations prevented us from fully exploring the effectiveness of the self-attention mechanism in argument structure learning. 
In future work, we might leverage more powerful GPUs when implementing a multi-head self-attention mechanism and thoroughly investigate its impact. Nevertheless the performance of our model is better than previous state-of-the-art models that use attention and are computationally more expensive. 
% \textbf{2)} 
% % Our experiments indicate that a larger window size can lead to performance improvements.
% Our method relies on relatively large computation resources, i.e., at least one GPU with 24 GB VRAM. 
% Experimental results are not optimal if we using one GPU with 12 GB VRAM. 
% Our experiments indicate that a larger window size can lead to performance improvements.
% As a result, we intend to conduct further testing of the DMON model without symmetry cropping to explore its potential benefits.
\textbf{2)} We will test whether our bidirectional learning mechanism can be embedded to other pairwise classification models in NLP to improve their performance. 
% \textbf{3)} We found that our model suffers from the label imbalance problem.
In the future, we will leverage balancing approaches, such as focal loss and resampling, to alleviate this problem. 

\section{Acknowledgements}
This research was funded by the CHIST-ERA projects ANTIDOTE (ERA-NET CHIST-ERA IV FET PROACT JTC 2019) and iTRUST (ERA-NET CHIST-ERA JTC 2021). 

\nocite{*}
\section{Bibliographical References}\label{sec:reference}

\bibliographystyle{lrec-coling2024-natbib}
\bibliography{lrec-coling2024-example}

\begin{thebibliography}{37}
\expandafter\ifx\csname natexlab\endcsname\relax\def\natexlab#1{#1}\fi

\bibitem[{Alzubaidi et~al.(2021)Alzubaidi, Zhang, Humaidi, Al-Dujaili, Duan, Al-Shamma, Santamar{\'\i}a, Fadhel, Al-Amidie, and Farhan}]{alzubaidi2021review}
Laith Alzubaidi, Jinglan Zhang, Amjad~J Humaidi, Ayad Al-Dujaili, Ye~Duan, Omran Al-Shamma, Jos{\'e} Santamar{\'\i}a, Mohammed~A Fadhel, Muthana Al-Amidie, and Laith Farhan. 2021.
\newblock Review of deep learning: Concepts, cnn architectures, challenges, applications, future directions.
\newblock \emph{Journal of big Data}, 8:1--74.

\bibitem[{Andreoli(2019)}]{andreoli2019convolution}
Jean-Marc Andreoli. 2019.
\newblock Convolution, attention and structure embedding.
\newblock \emph{arXiv preprint arXiv:1905.01289}.

\bibitem[{Bao et~al.(2021)Bao, Fan, Wu, Dang, Du, and Xu}]{bao2021neural}
Jianzhu Bao, Chuang Fan, Jipeng Wu, Yixue Dang, Jiachen Du, and Ruifeng Xu. 2021.
\newblock A neural transition-based model for argumentation mining.
\newblock In \emph{Proceedings of the 59th Annual Meeting of the Association for Computational Linguistics and the 11th International Joint Conference on Natural Language Processing (Volume 1: Long Papers)}, pages 6354--6364.

\bibitem[{Brown et~al.(2020)Brown, Mann, Ryder, Subbiah, Kaplan, Dhariwal, Neelakantan, Shyam, Sastry, Askell, Agarwal, Herbert-Voss, Krueger, Henighan, Child, Ramesh, Ziegler, Wu, Winter, Hesse, Chen, Sigler, Litwin, Gray, Chess, Clark, Berner, McCandlish, Radford, Sutskever, and Amodei}]{NEURIPS2020_1457c0d6}
Tom Brown, Benjamin Mann, Nick Ryder, Melanie Subbiah, Jared~D Kaplan, Prafulla Dhariwal, Arvind Neelakantan, Pranav Shyam, Girish Sastry, Amanda Askell, Sandhini Agarwal, Ariel Herbert-Voss, Gretchen Krueger, Tom Henighan, Rewon Child, Aditya Ramesh, Daniel Ziegler, Jeffrey Wu, Clemens Winter, Chris Hesse, Mark Chen, Eric Sigler, Mateusz Litwin, Scott Gray, Benjamin Chess, Jack Clark, Christopher Berner, Sam McCandlish, Alec Radford, Ilya Sutskever, and Dario Amodei. 2020.
\newblock \href {https://proceedings.neurips.cc/paper_files/paper/2020/file/1457c0d6bfcb4967418bfb8ac142f64a-Paper.pdf} {Language models are few-shot learners}.
\newblock In \emph{Advances in Neural Information Processing Systems}, volume~33, pages 1877--1901. Curran Associates, Inc.

\bibitem[{Devlin et~al.(2019)Devlin, Chang, Lee, and Toutanova}]{devlin2018bert}
Jacob Devlin, Ming-Wei Chang, Kenton Lee, and Kristina Toutanova. 2019.
\newblock \href {https://doi.org/10.18653/v1/N19-1423} {{BERT}: Pre-training of deep bidirectional transformers for language understanding}.
\newblock In \emph{Proceedings of the 2019 Conference of the North {A}merican Chapter of the Association for Computational Linguistics: Human Language Technologies, Volume 1 (Long and Short Papers)}, pages 4171--4186, Minneapolis, Minnesota. Association for Computational Linguistics.

\bibitem[{Ding et~al.(2022)Ding, Zhang, Han, and Ding}]{ding2022scaling}
Xiaohan Ding, Xiangyu Zhang, Jungong Han, and Guiguang Ding. 2022.
\newblock Scaling up your kernels to 31x31: Revisiting large kernel design in cnns.
\newblock In \emph{Proceedings of the IEEE/CVF Conference on Computer Vision and Pattern Recognition}, pages 11963--11975.

\bibitem[{Dumoulin and Visin(2016)}]{dumoulin2016guide}
Vincent Dumoulin and Francesco Visin. 2016.
\newblock A guide to convolution arithmetic for deep learning.
\newblock \emph{arXiv preprint arXiv:1603.07285}.

\bibitem[{Eger et~al.(2017)Eger, Daxenberger, and Gurevych}]{eger2017neural}
Steffen Eger, Johannes Daxenberger, and Iryna Gurevych. 2017.
\newblock Neural end-to-end learning for computational argumentation mining.
\newblock \emph{arXiv preprint arXiv:1704.06104}.

\bibitem[{Ein-Dor et~al.(2022)Ein-Dor, Shnayderman, Spector, Dankin, Aharonov, and Slonim}]{ein2022fortunately}
Liat Ein-Dor, Ilya Shnayderman, Artem Spector, Lena Dankin, Ranit Aharonov, and Noam Slonim. 2022.
\newblock Fortunately, discourse markers can enhance language models for sentiment analysis.
\newblock In \emph{Proceedings of the AAAI Conference on Artificial Intelligence}, volume~36, pages 10608--10617.

\bibitem[{Galassi et~al.(2018)Galassi, Lippi, and Torroni}]{galassi2018argumentative}
Andrea Galassi, Marco Lippi, and Paolo Torroni. 2018.
\newblock Argumentative link prediction using residual networks and multi-objective learning.
\newblock In \emph{Proceedings of the 5th Workshop on Argument Mining}, pages 1--10.

\bibitem[{Galassi et~al.(2020)Galassi, Lippi, and Torroni}]{galassi2020attention}
Andrea Galassi, Marco Lippi, and Paolo Torroni. 2020.
\newblock Attention in natural language processing.
\newblock \emph{IEEE transactions on neural networks and learning systems}, 32(10):4291--4308.

\bibitem[{Galassi et~al.(2021)Galassi, Lippi, and Torroni}]{galassi2021multi}
Andrea Galassi, Marco Lippi, and Paolo Torroni. 2021.
\newblock Multi-task attentive residual networks for argument mining.
\newblock \emph{IEEE/ACM Transactions on Audio, Speech, and Language Processing}.

\bibitem[{Hua and Wang(2022)}]{hua2022efficient}
Xinyu Hua and Lu~Wang. 2022.
\newblock \href {https://doi.org/10.18653/v1/2022.findings-acl.36} {Efficient argument structure extraction with transfer learning and active learning}.
\newblock In \emph{Findings of the Association for Computational Linguistics: ACL 2022}, pages 423--437, Dublin, Ireland. Association for Computational Linguistics.

\bibitem[{Imrich et~al.(2008)Imrich, Klavzar, and Rall}]{imrich2008topics}
Wilfried Imrich, Sandi Klavzar, and Douglas~F Rall. 2008.
\newblock \emph{Topics in graph theory: Graphs and their Cartesian product}.
\newblock CRC Press.

\bibitem[{Jiang et~al.(2023)Jiang, Liu, Ma, Zhang, Sachan, and Cotterell}]{jiang2023discourse}
Yuchen~Eleanor Jiang, Tianyu Liu, Shuming Ma, Dongdong Zhang, Mrinmaya Sachan, and Ryan Cotterell. 2023.
\newblock Discourse centric evaluation of machine translation with a densely annotated parallel corpus.
\newblock \emph{arXiv preprint arXiv:2305.11142}.

\bibitem[{Lawrence and Reed(2020)}]{lawrence2020argument}
John Lawrence and Chris Reed. 2020.
\newblock Argument mining: A survey.
\newblock \emph{Computational Linguistics}, 45(4):765--818.

\bibitem[{Li and Yu(2020)}]{li2020icd}
Fei Li and Hong Yu. 2020.
\newblock Icd coding from clinical text using multi-filter residual convolutional neural network.
\newblock In \emph{proceedings of the AAAI conference on artificial intelligence}, volume~34, pages 8180--8187.

\bibitem[{Li et~al.(2018)Li, Fang, Mei, and Zhang}]{li2018multi}
Juncheng Li, Faming Fang, Kangfu Mei, and Guixu Zhang. 2018.
\newblock Multi-scale residual network for image super-resolution.
\newblock In \emph{Proceedings of the European conference on computer vision (ECCV)}, pages 517--532.

\bibitem[{Liu et~al.(2019)Liu, Ott, Goyal, Du, Joshi, Chen, Levy, Lewis, Zettlemoyer, and Stoyanov}]{liu2019roberta}
Yinhan Liu, Myle Ott, Naman Goyal, Jingfei Du, Mandar Joshi, Danqi Chen, Omer Levy, Mike Lewis, Luke Zettlemoyer, and Veselin Stoyanov. 2019.
\newblock Roberta: A robustly optimized bert pretraining approach.
\newblock \emph{arXiv preprint arXiv:1907.11692}.

\bibitem[{Loshchilov and Hutter(2019)}]{loshchilov2017decoupled}
Ilya Loshchilov and Frank Hutter. 2019.
\newblock \href {https://openreview.net/forum?id=Bkg6RiCqY7} {Decoupled weight decay regularization}.
\newblock In \emph{International Conference on Learning Representations}.

\bibitem[{Luo et~al.(2016)Luo, Li, Urtasun, and Zemel}]{luo2016understanding}
Wenjie Luo, Yujia Li, Raquel Urtasun, and Richard Zemel. 2016.
\newblock Understanding the effective receptive field in deep convolutional neural networks.
\newblock \emph{Advances in neural information processing systems}, 29.

\bibitem[{Mayer et~al.(2020)Mayer, Cabrio, and Villata}]{mayer2020transformer}
Tobias Mayer, Elena Cabrio, and Serena Villata. 2020.
\newblock Transformer-based argument mining for healthcare applications.
\newblock In \emph{ECAI 2020}, pages 2108--2115. IOS Press.

\bibitem[{Mochales and Moens(2011)}]{mochales2011argumentation}
Raquel Mochales and Marie-Francine Moens. 2011.
\newblock Argumentation mining.
\newblock \emph{Artificial Intelligence and Law}, 19:1--22.

\bibitem[{Moens(2013)}]{Moens2013}
Marie{-}Francine Moens. 2013.
\newblock \href {https://doi.org/10.1145/2701336.2701635} {Argumentation mining: Where are we now, where do we want to be and how do we get there?}
\newblock In \emph{Proceedings of the 5th 2013 Forum on Information Retrieval Evaluation, {FIRE} '13, New Delhi, India, December 4-6, 2013}, pages 2:1--2:6. {ACM}.

\bibitem[{Morio et~al.(2020)Morio, Ozaki, Morishita, Koreeda, and Yanai}]{morio2020towards}
Gaku Morio, Hiroaki Ozaki, Terufumi Morishita, Yuta Koreeda, and Kohsuke Yanai. 2020.
\newblock Towards better non-tree argument mining: Proposition-level biaffine parsing with task-specific parameterization.
\newblock In \emph{Proceedings of the 58th Annual Meeting of the Association for Computational Linguistics}, pages 3259--3266.

\bibitem[{Niculae et~al.(2017)Niculae, Park, and Cardie}]{niculae2017argument}
Vlad Niculae, Joonsuk Park, and Claire Cardie. 2017.
\newblock \href {https://doi.org/10.18653/v1/P17-1091} {Argument mining with structured {SVM}s and {RNN}s}.
\newblock In \emph{Proceedings of the 55th Annual Meeting of the Association for Computational Linguistics (Volume 1: Long Papers)}, pages 985--995, Vancouver, Canada. Association for Computational Linguistics.

\bibitem[{Palau and Moens(2009)}]{palau2009argumentation}
Raquel~Mochales Palau and Marie-Francine Moens. 2009.
\newblock Argumentation mining: the detection, classification and structure of arguments in text.
\newblock In \emph{Proceedings of the 12th international conference on artificial intelligence and law}, pages 98--107.

\bibitem[{Poudyal et~al.(2020)Poudyal, {\v{S}}avelka, Ieven, Moens, Goncalves, and Quaresma}]{poudyal2020echr}
Prakash Poudyal, Jarom{\'\i}r {\v{S}}avelka, Aagje Ieven, Marie~Francine Moens, Teresa Goncalves, and Paulo Quaresma. 2020.
\newblock Echr: legal corpus for argument mining.
\newblock In \emph{Proceedings of the 7th Workshop on Argument Mining}, pages 67--75.

\bibitem[{Stylianou and Vlahavas(2021)}]{stylianou2021transformed}
Nikolaos Stylianou and Ioannis Vlahavas. 2021.
\newblock Transformed: End-to-end transformers for evidence-based medicine and argument mining in medical literature.
\newblock \emph{Journal of Biomedical Informatics}, 117:103767.

\bibitem[{Vyas et~al.(2018)Vyas, Jammalamadaka, Zhu, Das, Kaul, and Willke}]{vyas2018out}
Apoorv Vyas, Nataraj Jammalamadaka, Xia Zhu, Dipankar Das, Bharat Kaul, and Theodore~L Willke. 2018.
\newblock Out-of-distribution detection using an ensemble of self supervised leave-out classifiers.
\newblock In \emph{Proceedings of the European Conference on Computer Vision (ECCV)}, pages 550--564.

\bibitem[{Weng et~al.(2023)Weng, Luo, Lin, Li, and Zhong}]{weng2023boosting}
Juanjuan Weng, Zhiming Luo, Dazhen Lin, Shaozi Li, and Zhun Zhong. 2023.
\newblock Boosting adversarial transferability via fusing logits of top-1 decomposed feature.
\newblock \emph{arXiv preprint arXiv:2305.01361}.

\bibitem[{Wolf and Gibson(2006)}]{wolf2006coherence}
Florian Wolf and Edward Gibson. 2006.
\newblock \emph{Coherence in natural language: data structures and applications}.
\newblock MIT Press.

\bibitem[{Xiong et~al.(2019)Xiong, He, Wu, and Wang}]{xiong2019modeling}
Hao Xiong, Zhongjun He, Hua Wu, and Haifeng Wang. 2019.
\newblock Modeling coherence for discourse neural machine translation.
\newblock In \emph{Proceedings of the AAAI conference on artificial intelligence}, volume~33, pages 7338--7345.

\bibitem[{Xu et~al.(2020)Xu, Gan, Cheng, and Liu}]{xu2019discourse}
Jiacheng Xu, Zhe Gan, Yu~Cheng, and Jingjing Liu. 2020.
\newblock \href {https://doi.org/10.18653/v1/2020.acl-main.451} {Discourse-aware neural extractive text summarization}.
\newblock In \emph{Proceedings of the 58th Annual Meeting of the Association for Computational Linguistics}, pages 5021--5031, Online. Association for Computational Linguistics.

\bibitem[{Yang and Li(2018)}]{yang2018scidtb}
An~Yang and Sujian Li. 2018.
\newblock \href {https://doi.org/10.18653/v1/P18-2071} {{S}ci{DTB}: Discourse dependency {T}ree{B}ank for scientific abstracts}.
\newblock In \emph{Proceedings of the 56th Annual Meeting of the Association for Computational Linguistics (Volume 2: Short Papers)}, pages 444--449, Melbourne, Australia. Association for Computational Linguistics.

\bibitem[{Yasunaga et~al.(2022)Yasunaga, Leskovec, and Liang}]{yasunaga2022linkbert}
Michihiro Yasunaga, Jure Leskovec, and Percy Liang. 2022.
\newblock \href {https://doi.org/10.18653/v1/2022.acl-long.551} {{L}ink{BERT}: Pretraining language models with document links}.
\newblock In \emph{Proceedings of the 60th Annual Meeting of the Association for Computational Linguistics (Volume 1: Long Papers)}, pages 8003--8016, Dublin, Ireland. Association for Computational Linguistics.

\bibitem[{Zhang et~al.(2022)Zhang, Hu, and Wang}]{zhang2022parc}
Haokui Zhang, Wenze Hu, and Xiaoyu Wang. 2022.
\newblock Parc-net: Position aware circular convolution with merits from convnets and transformer.
\newblock In \emph{Computer Vision--ECCV 2022: 17th European Conference, Tel Aviv, Israel, October 23--27, 2022, Proceedings, Part XXVI}, pages 613--630. Springer.

\end{thebibliography}

\end{document}